\documentclass{article}

\usepackage{times}
\usepackage{graphicx} % more modern
\usepackage{subfigure} 

\usepackage{natbib}

\usepackage{algorithm}
\usepackage{algorithmic}

\usepackage{hyperref}

\usepackage{amsmath}
\usepackage{amsfonts}
\usepackage{amssymb}

\newcommand{\figspace}{\vspace{.155cm}}
\newcommand{\figwidth}{0.454\textwidth}
\newcommand{\tbl}[1]{Table~\ref{#1}}

\newcommand{\sigmoid}{\boldsymbol{\sigma}}
\newcommand{\linear}{\mathit{W}}
\newcommand{\enc}{\mathit{enc}}
\newcommand{\dec}{\mathit{dec}}
\newcommand{\hdec}{h^{\dec}}
\newcommand{\henc}{h^{\enc}}
\newcommand{\encoder}{\mathit{RNN}^{\enc}}
\newcommand{\decoder}{\mathit{RNN}^{\dec}}
\newcommand{\readop}{\mathit{read}}
\newcommand{\writeop}{\mathit{write}}

\newcommand{\canv}{c}

\newcommand{\post}{Q}
\newcommand{\prior}{P}
\newcommand{\lat}{z}
\newcommand{\Lat}{Z}
\newcommand{\mean}{\mu}
\newcommand{\std}{\sigma}
\newcommand{\gauss}{\mathcal{N}}
\newcommand{\loss}{\mathcal{L}}
\newcommand{\lrec}{\loss^{x}}
\newcommand{\llat}{\loss^{z}}

\newcommand{\data}{D}
\newcommand{\kl}[2]{\mathit{KL}\big(#1 || #2\big)}

\newcommand*\pd[3][]
{ 
\def\next{#1}%
\ifx\empty\next
  \frac{\partial#2}{\partial #3}
\else
  \frac{{\partial^{#1} #2}}{\partial#3^{#1}}
\fi
}
\newcommand{\expect}[3]{\left\langle #1 \right\rangle_{#2 \sim #3}}
\usepackage[accepted]{icml2015}

\icmltitlerunning{DRAW: A Recurrent Neural Network For Image Generation}

\begin{document} 

\twocolumn[
\icmltitle{DRAW: A Recurrent Neural Network For Image Generation}

% unless you've provided the [accepted] option to the icml2015
\icmlauthor{Karol Gregor}{karolg@google.com}
\icmlauthor{Ivo Danihelka}{danihelka@google.com}
\icmlauthor{Alex Graves}{gravesa@google.com}
\icmlauthor{Danilo Jimenez Rezende}{danilor@google.com}
\icmlauthor{Daan Wierstra}{wierstra@google.com}
\icmladdress{Google DeepMind}

\icmlkeywords{generative models, deep learning, variational inference, recurrent neural networks}

\vskip 0.3in
]

\begin{abstract}
This paper introduces the \emph{Deep Recurrent Attentive Writer} (DRAW) neural network architecture for image generation. 
DRAW networks combine a novel spatial attention mechanism that mimics the foveation of the human eye, with a sequential variational auto-encoding framework that allows for the iterative construction of complex images.
The system substantially improves on the state of the art for generative models on MNIST, and, when trained on the Street View House Numbers dataset, it generates images that cannot be distinguished from real data with the naked eye.
\end{abstract}

\section{Introduction}
A person asked to draw, paint or otherwise recreate a visual scene will naturally do so in a sequential, iterative fashion, reassessing their handiwork after each modification.
Rough outlines are gradually replaced by precise forms, lines are sharpened, darkened or erased, shapes are altered, and the final picture emerges.
Most approaches to automatic image generation, however, aim to generate entire scenes at once.
In the context of generative neural networks, this typically means that all the pixels are conditioned on a single latent distribution~\citep{dayan1995helmholtz,hinton2006reducing,larochelle2011neural}.
As well as precluding the possibility of iterative self-correction, the ``one shot'' approach is fundamentally difficult to scale to large images.
The \emph{Deep Recurrent Attentive Writer} (DRAW) architecture represents a shift towards a more natural form of image construction, in which parts of a scene are created independently from others, and approximate sketches are successively refined.

\begin{figure}[t]
% \vspace{-.0cm}
\begin{center}
\begin{minipage}{\figwidth}
\includegraphics[width=0.99\textwidth]{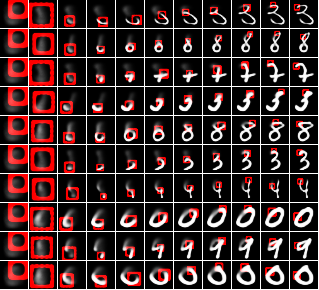}
\includegraphics[width=0.18\linewidth]{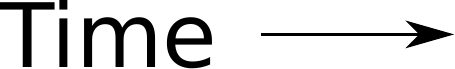}
\end{minipage}
\end{center}
\caption{\textbf{A trained DRAW network generating MNIST digits.}
Each row shows successive stages in the generation of a single digit.
Note how the lines composing the digits appear to be ``drawn'' by the network. 
The red rectangle delimits the area attended to by the network at each time-step, 
with the focal precision indicated by the width of the rectangle border. 
}
\label{fig:mnist_generative_process}
% \vspace{-.4cm}
\end{figure}

The core of the DRAW architecture is a pair of recurrent neural networks: an \emph{encoder} network that compresses the real images presented during training, and a \emph{decoder} that reconstitutes images after receiving codes.
The combined system is trained end-to-end with stochastic gradient descent, where the loss function is a variational upper bound on the log-likelihood of the data.
It therefore belongs to the family of \emph{variational auto-encoders}, a recently emerged hybrid of deep learning and variational inference that has led to significant advances in generative modelling~\citep{gregor2013deep,kingma2013auto,rezende2014stochastic,mnih2014nvil,salimans2014markov}.
Where DRAW differs from its siblings is that, rather than generating images in a single pass, it iteratively constructs scenes through an accumulation of modifications emitted by the decoder, each of which is observed by the encoder.

An obvious correlate of generating images step by step is the ability to selectively attend to parts of the scene while ignoring others.
A wealth of results in the past few years suggest that visual structure can be better captured by a sequence of partial glimpses, or foveations, than by a single sweep through the entire image \citep{larochelle2010glimpses,denil2012learning,tang2013learning,ranzato2014learning,zheng2014neural,mnih2014recurrent,ba2014multiple,sermanet2014attention}.
The main challenge faced by sequential attention models is learning where to look, which can be addressed with reinforcement learning techniques such as policy gradients~\citep{mnih2014recurrent}.
The attention model in DRAW, however, is fully differentiable, making it possible to train with standard backpropagation.
In this sense it resembles the selective read and write operations developed for the Neural Turing Machine~\citep{graves2014neural}.

The following section defines the DRAW architecture, along with the loss function used for training and the procedure for image generation.
Section~\ref{sec:readwrite} presents the selective attention model and shows how it is applied to reading and modifying images.
Section~\ref{sec:results} provides experimental results on the MNIST, Street View House Numbers and CIFAR-10 datasets, with examples of generated images; and concluding remarks are given in Section~\ref{sec:conclusion}.
Lastly, we would like to direct the reader to the video accompanying this paper (\url{https://www.youtube.com/watch?v=Zt-7MI9eKEo}) which contains examples of DRAW networks reading and generating images.

\section{The DRAW Network}
\label{sec:ThinkCore}

\begin{figure}[t]
\begin{center}
\includegraphics[width=\figwidth]{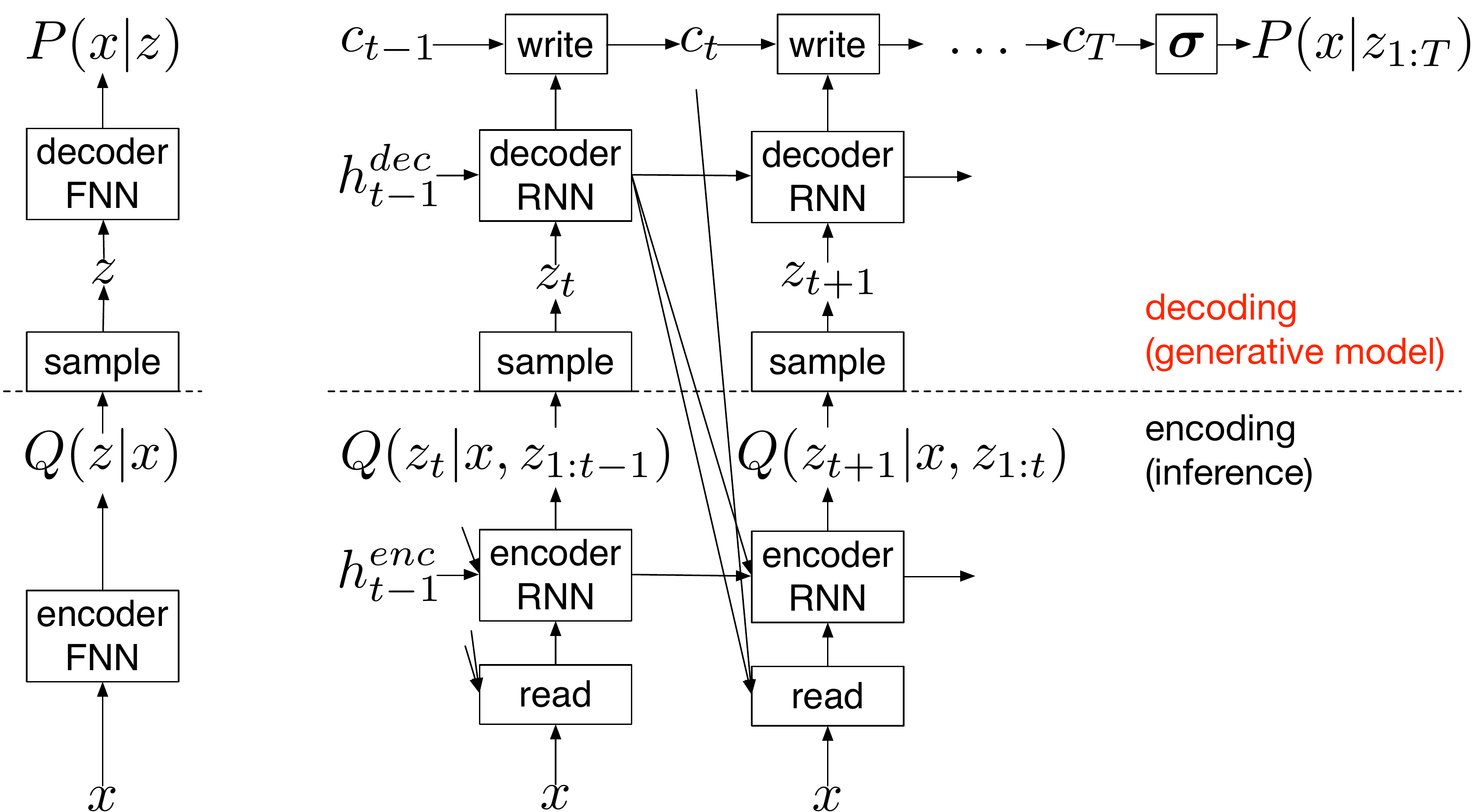}
\end{center}
\caption{\textbf{Left: Conventional Variational Auto-Encoder}. During generation, a sample $z$ is drawn from a prior $\prior(z)$ and passed through the feedforward decoder network to compute the probability of the input $\prior(x|z)$ given the sample. During inference the input $x$ is passed to the encoder network, producing an approximate posterior $\post(z|x)$ over latent variables. During training, $z$ is sampled from $\post(z|x)$ and then used to compute the total description length $\kl{\post(\Lat|x)}{\prior(\Lat)}{} - \log(\prior(x|\lat))$, which is minimised with stochastic gradient descent.
\textbf{Right: DRAW Network}. 
At each time-step a sample $z_t$ from the prior $\prior(z_t)$ is passed to the recurrent decoder network, which then modifies part of the canvas matrix. The final canvas matrix $\canv_T$ is used to compute $\prior(x|z_{1:T})$. During inference the input is read at every time-step and the result is passed to the encoder RNN. The RNNs at the previous time-step specify where to read. The output of the encoder RNN is used to compute the approximate posterior over the latent variables at that time-step. 
}
\label{fig:diagram}
\end{figure}

The basic structure of a DRAW network is similar to that of other variational auto-encoders: an encoder network determines a distribution over latent codes that capture salient information about the input data; a decoder network receives samples from the code distribuion and uses them to condition its own distribution over images.
However there are three key differences.
Firstly, both the encoder and decoder are recurrent networks in DRAW, so that a \emph{sequence} of code samples is exchanged between them; moreover the encoder is privy to the decoder's previous outputs, allowing it to tailor the codes it sends according to the decoder's behaviour so far. 
Secondly, the decoder's outputs are successively added to the distribution that will ultimately generate the data, as opposed to emitting this distribution in a single step.
And thirdly, a dynamically updated attention mechanism is used to restrict both the input region observed by the encoder, and the output region modified by the decoder.
In simple terms, the network decides at each time-step ``where to read'' and ``where to write'' as well as ``what to write''. 
The architecture is sketched in Fig.~\ref{fig:diagram}, alongside a feedforward variational auto-encoder. 

\subsection{Network Architecture}

Let $\encoder$ be the function enacted by the encoder network at a single time-step.
The output of $\encoder$ at time $t$ is the encoder hidden vector $\henc_t$.
Similarly the output of the decoder $\decoder$ at $t$ is the hidden vector $\hdec_t$.
In general the encoder and decoder may be implemented by any recurrent neural network.
In our experiments we use the \emph{Long Short-Term Memory} architecture (LSTM;~\citet{hochreiter1997long}) for both, in the extended form with \emph{forget gates}~\citep{gers2000learning}.
We favour LSTM due to its proven track record for handling long-range dependencies in real sequential data~\citep{graves2013generating,sutskever2014sequence}.
Throughout the paper, we use the notation $b = \linear(a)$ to denote a linear weight matrix with bias from the vector $a$ to the vector $b$.

At each time-step $t$, the encoder receives input from both the image $x$ and from the previous decoder hidden vector $\hdec_{t-1}$.
The precise form of the encoder input depends on a $\readop$ operation, which will be defined in the next section.
The output $\henc_t$ of the encoder is used to parameterise a distribution $\post(\Lat_t|\henc_t)$ over the latent vector $\lat_t$.
In our experiments the latent distribution is a diagonal Gaussian $\gauss(\Lat_t|\mean_t,\std_t)$:
\begin{align}
\label{eq:mu}
\mean_t &= \linear(h^{enc}_t)\\
\label{eq:sigma}
\std_t &= \exp\left(\linear(h^{enc}_t)\right)
\end{align}
Bernoulli distributions are more common than Gaussians for latent variables in auto-encoders \citep{dayan1995helmholtz,gregor2013deep}; however a great advantage of Gaussian latents is that the gradient of a function of the samples with respect to the distribution parameters can be easily obtained using the so-called \emph{reparameterization trick} \citep{kingma2013auto,rezende2014stochastic}.
This makes it straightforward to back-propagate unbiased, low variance stochastic gradients of the loss function through the latent distribution.

At each time-step a sample $\lat_t \sim \post(\Lat_t|\henc_t)$ drawn from the latent distribution is passed as input to the decoder.
The output $\hdec_t$ of the decoder is added (via a $\writeop$ operation, defined in the sequel) to a cumulative \emph{canvas} matrix $\canv_t$, which is ultimately used to reconstruct the image.
The total number of time-steps $T$ consumed by the network before performing the reconstruction is a free parameter that must be specified in advance.

For each image $x$ presented to the network, $\canv_0, \henc_0, \hdec_0$ are initialised to learned biases, and the DRAW network iteratively computes the following equations for $t=1\dots,T$:
\begin{align}
\label{eq:x_hat}
\hat{x}_t &= x-\sigmoid(\canv_{t-1})\\
\label{eq:read}
r_t &= \readop(x_t, \hat{x}_t, \hdec_{t-1})\\
\henc_t &= \encoder(\henc_{t-1}, [r_t, \hdec_{t-1}])\\
\lat_t &\sim \post(\Lat_t|\henc_t)\\
\hdec_t &= \decoder(\hdec_{t-1}, z_t)\\
\label{eq:write}
\canv_t &= \canv_{t-1} + \writeop(\hdec_t)
\end{align}
where $\hat{x}_t$ is the \emph{error image}, $[v,w]$ is the concatenation of vectors $v$ and $w$ into a single vector, and $\sigmoid$ denotes the logistic sigmoid function: $\sigmoid(x)=\frac{1}{1+\exp(-x)}$.
Note that $\henc_t$, and hence $\post(\Lat_t|\henc_t)$, depends on both $x$ and the history $\lat_{1:t-1}$ of previous latent samples.
We will sometimes make this dependency explicit by writing $\post(\Lat_t|x,\lat_{1:t-1})$, as shown in Fig.~\ref{fig:diagram}.
$\henc$ can also be passed as input to the $read$ operation; however we did not find that this helped performance and therefore omitted it.

\subsection{Loss Function}\label{sec:loss}

The final canvas matrix $\canv_T$ is used to parameterise a model $\data(X|\canv_T)$ of the input data.
If the input is binary, the natural choice for $\data$ is a Bernoulli distribution with means given by $\sigmoid(\canv_T)$.
The \emph{reconstruction loss} $\lrec$ is defined as the negative log probability of $x$ under $\data$:
\begin{equation}
\label{eq:reconstruction}
\lrec = -\log \data(x|\canv_T)
\end{equation}
The \emph{latent loss} $\llat$ for a sequence of latent distributions $\post(\Lat_t|\henc_t)$ is defined as the summed Kullback-Leibler divergence of some latent prior $\prior(\Lat_t)$ from $\post(\Lat_t|\henc_t)$:
\begin{equation}
\label{eq:kl}
\llat = \sum_{t=1}^T{\kl{\post(\Lat_t|\henc_t)}{\prior(\Lat_t)}}
\end{equation}
Note that this loss depends upon the latent samples $\lat_t$ drawn from $\post(\Lat_t|\henc_t)$, which depend in turn on the input $x$.
If the latent distribution is a diagonal Gaussian with $\mean_t$, $\sigma_t$ as defined in Eqs~\ref{eq:mu} and~\ref{eq:sigma}, a simple choice for $\prior(\Lat_t)$ is a standard Gaussian with mean zero and standard deviation one, in which case Eq.~\ref{eq:kl} becomes
\begin{equation}
\llat = \frac{1}{2}\left(\sum_{t=1}^T{\mean_t^2 + \std_t^2 - \log \std_t^2}\right) - T/2
\end{equation}
The total loss $\loss$ for the network is the expectation of the sum of the reconstruction and latent losses:
\begin{equation}
\label{eq:loss}
\loss = \expect{\lrec + \llat}{z}{\post}
\end{equation} 
which we optimise using a single sample of $z$ for each stochastic gradient descent step.

$\llat$ can be interpreted as the number of nats required to transmit the latent sample sequence $\lat_{1:T}$ to the decoder from the prior, and (if $x$ is discrete) $\lrec$ is the number of nats required for the decoder to reconstruct $x$ given $\lat_{1:T}$.
The total loss is therefore equivalent to the expected compression of the data by the decoder and prior.

\subsection{Stochastic Data Generation}\label{sec:generation}

An image $\tilde{x}$ can be generated by a DRAW network by iteratively picking latent samples $\tilde{\lat}_t$ from the prior $\prior$, then running the decoder to update the canvas matrix $\tilde{c}_t$.
After $T$ repetitions of this process the generated image is a sample from $\data(X|\tilde{c}_T)$:
\begin{align}
\tilde{\lat}_t &\sim \prior(\Lat_t)\\
\tilde{h}^{dec}_t &= \decoder(\tilde{h}^{dec}_{t-1}, \tilde{\lat}_t)\\
\tilde{\canv}_t &= \tilde{\canv}_{t-1} + write(\tilde{h}^{dec}_t)\\
\tilde{x} &\sim \data(X|\tilde{\canv_T})
\end{align}
Note that the encoder is not involved in image generation.

\section{Read and Write Operations}
\label{sec:readwrite}
The DRAW network described in the previous section is not complete until the $\readop$ and $\writeop$ operations in Eqs.~\ref{eq:read} and~\ref{eq:write} have been defined.
This section describes two ways to do so, one with selective attention and one without.

\subsection{Reading and Writing Without Attention}\label{sec:noattention}
In the simplest instantiation of DRAW the entire input image is passed to the encoder at every time-step, and the decoder modifies the entire canvas matrix at every time-step.
In this case the $\readop$ and $\writeop$ operations reduce to
\begin{align}
\label{eq:linearread}
\readop(x, \hat{x}_t, \hdec_{t-1}) &= [x, \hat{x}_t]\\
\label{eq:linearwrite}
\writeop(\hdec_t) &= \linear(\hdec_t)
\end{align}
However this approach does not allow the encoder to focus on only part of the input when creating the latent distribution; nor does it allow the decoder to modify only a part of the canvas vector.
In other words it does not provide the network with an explicit selective attention mechanism, which we believe to be crucial to large scale image generation.
We refer to the above configuration as ``DRAW without attention''.

\subsection{Selective Attention Model}\label{sec:attention}
To endow the network with selective attention without sacrificing the benefits of gradient descent training, we take inspiration from the differentiable attention mechanisms recently used in handwriting synthesis~\citep{graves2013generating} and Neural Turing Machines~\citep{graves2014neural}.
Unlike the aforementioned works, we consider an explicitly two-dimensional form of attention, where an array of 2D Gaussian filters is applied to the image, yielding an image `patch' of smoothly varying location and zoom.
This configuration, which we refer to simply as ``DRAW'', somewhat resembles the affine transformations used in computer graphics-based autoencoders~\cite{tieleman2014optimizing}.

\begin{figure}[t]
\begin{center}
\includegraphics[width=.45\textwidth]{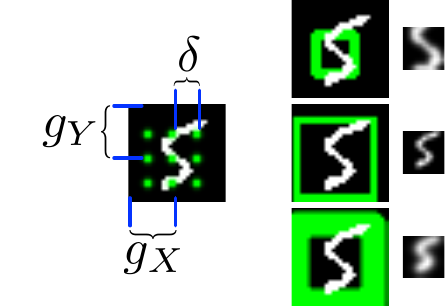}
\end{center}
\caption{\textbf{Left:} A $3 \times 3$ grid of filters superimposed on an image. The stride ($\delta$) and centre location ($g_X,g_Y$) are indicated.
\textbf{Right:} Three $N \times N$ patches extracted from the image ($N = 12$). The green rectangles on the left indicate the boundary and precision ($\sigma$) of the patches, while the patches themselves are shown to the right. The top patch has a small $\delta$ and high $\sigma$, giving a zoomed-in but blurry view of the centre of the digit; the middle patch has large $\delta$ and low $\sigma$, effectively downsampling the whole image; and the bottom patch has high $\delta$ and $\sigma$.
}
\label{fig:marked_grid}
\end{figure}

As illustrated in Fig.~\ref{fig:marked_grid}, the $N \times N$ grid of Gaussian filters is positioned on the image by specifying the co-ordinates of the grid centre and the stride distance between adjacent filters.
The stride controls the `zoom' of the patch;
that is, the larger the stride, the larger an area of the original image will be visible in the attention patch, but the lower the effective resolution of the patch will be.
The grid centre $(g_X,g_Y)$ and stride $\delta$ (both of which are real-valued) determine the mean location $\mu_X^i, \mu_Y^j$ of the filter at row $i$, column $j$ in the patch as follows:
\begin{align}
\mu_X^i  &= g_X + \left(i-N/2-0.5\right)\delta\\
\mu_Y^j  &= g_Y + \left(j-N/2-0.5\right)\delta
\end{align}
Two more parameters are required to fully specify the attention model: the isotropic variance $\sigma^2$ of the Gaussian filters, and a scalar intensity $\gamma$ that multiplies the filter response.
Given an $A \times B$ input image $x$,
all five attention parameters are dynamically determined at each time step via a linear transformation of the decoder output $\hdec$:
\begin{align}
&(\tilde{g}_X, \tilde{g}_Y, \log \sigma^2,\log \tilde{\delta},\log \gamma) = \linear(\hdec)\\
&g_X = \frac{A + 1}{2}(\tilde{g}_X + 1)\\
&g_Y = \frac{B + 1}{2}(\tilde{g}_Y + 1)\\
&\delta = \frac{\max(A,B) - 1}{N - 1}\tilde{\delta}
\end{align}
where the variance, stride and intensity are emitted in the log-scale to ensure positivity. 
The scaling of $g_X$, $g_Y$ and $\delta$ is chosen to ensure that the initial patch (with a randomly initialised network) roughly covers the whole input image.

Given the attention parameters emitted by the decoder, the horizontal and vertical filterbank matrices $F_X$ and $F_Y$ (dimensions $N \times A$ and $N \times B$ respectively) are defined as follows:
\begin{align}
\label{eq:focus_x}
F_X[i, a] &= \frac{1}{Z_X}\exp\left(-\frac{(a - \mu_X^i)^2}{2\sigma^2}\right)\\
\label{eq:focus_y}
F_Y[j, b] &= \frac{1}{Z_Y}\exp\left(-\frac{(b - \mu_Y^j)^2}{2\sigma^2}\right)
\end{align}
where $(i,j)$ is a point in the attention patch, $(a,b)$ is a point in the input image, and $Z_x, Z_y$ are normalisation constants that ensure that $\sum_{a}F_X[i, a] = 1$ and $\sum_{b}F_Y[j, b] = 1$.

\begin{figure}[t]
\begin{center}
\figspace
\begin{minipage}{0.22\textwidth}
\includegraphics[width=.99\textwidth]{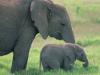}
\end{minipage}
\begin{minipage}{0.025\textwidth}
\includegraphics[width=.99\textwidth]{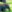}
\end{minipage}
\begin{minipage}{0.22\textwidth}
\includegraphics[width=.99\textwidth]{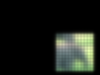}
\end{minipage}

\vspace{.05cm}
\begin{minipage}{0.22\textwidth}
\includegraphics[width=.99\textwidth]{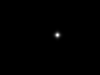}
\end{minipage}
\begin{minipage}{0.22\textwidth}
\includegraphics[width=.99\textwidth]{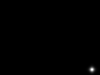}
\end{minipage}

\end{center}
\caption{\textbf{Zooming.} \textbf{Top Left:} The original $100 \times 75$ image.
\textbf{Top Middle:} A $12 \times 12$ patch extracted with 144 2D Gaussian filters.
\textbf{Top Right:} The reconstructed image when applying transposed filters on the patch.
\textbf{Bottom:} Only two 2D Gaussian filters are displayed. The first one is used to produce the top-left patch feature. The last filter is used to produce the bottom-right patch feature. By using different filter weights, the attention can be moved to a different location.
}
\label{fig:zooming}
\end{figure}

\subsection{Reading and Writing With Attention}

Given $F_X$, $F_Y$ and intensity $\gamma$ determined by $\hdec_{t-1}$, along with an input image $x$ and error image $\hat{x}_t$, the \emph{read} operation returns the concatenation of two $N \times N$ patches from the image and error image:
\begin{align}
\readop(x, \hat{x}_t, \hdec_{t-1}) &= \gamma[F_YxF_X^{T}, F_Y\hat{x}F_X^{T}]
\end{align}
Note that the same filterbanks are used for both the image and error image.
For the write operation, a distinct set of attention parameters $\hat{\gamma}$, $\hat{F}_X$ and $\hat{F}_Y$ are extracted from $\hdec_{t}$, the order of transposition is reversed, and the intensity is inverted:
\begin{align}
w_t &= \linear(\hdec_t)\\
\writeop(\hdec_{t}) &= \frac{1}{\hat{\gamma}} \hat{F}_Y^{T} w_t \hat{F}_X
\end{align}
where $w_t$ is the $N \times N$ \emph{writing patch} emitted by $\hdec_t$.
For colour images each point in the input and error image (and hence in the reading and writing patches) is an RGB triple. 
In this case the same reading and writing filters are used for all three channels.

\section{Experimental Results}\label{sec:results}

We assess the ability of DRAW to generate realistic-looking images by training on three datasets of progressively increasing visual complexity: MNIST~\cite{lecun1998gradient}, Street View House Numbers (SVHN)~\cite{netzer2011reading} and CIFAR-10~\cite{krizhevsky2009learning}.
The images generated by the network are always novel (not simply copies of training examples), and are virtually indistinguishable from real data for MNIST and SVHN; the generated CIFAR images are somewhat blurry, but still contain recognisable structure from natural scenes.
The binarized MNIST results substantially improve on the state of the art.
As a preliminary exercise, we also evaluate the 2D attention module of the DRAW network on cluttered MNIST classification.

For all experiments, the model $D(X|c_T)$ of the input data was a Bernoulli distribution with means given by $\sigmoid(c_T)$.
For the MNIST experiments, the reconstruction loss from Eq~\ref{eq:reconstruction} was the usual binary cross-entropy term.
For the SVHN and CIFAR-10 experiments, the red, green and blue pixel intensities were represented as numbers between 0 and 1, which were then interpreted as independent colour emission probabilities.
The reconstruction loss was therefore the cross-entropy between
the pixel intensities and the model probabilities.
Although this approach worked well in practice, it means that the training loss did not correspond to the true compression cost of RGB images.

Network hyper-parameters for all the experiments are presented in \tbl{tab:netdetails}.
The Adam optimisation algorithm~\citep{kingma2014adam} was used throughout.
Examples of generation sequences for MNIST and SVHN are provided in the accompanying video (\url{https://www.youtube.com/watch?v=Zt-7MI9eKEo}).

\subsection{Cluttered MNIST Classification}
\begin{figure}[t!]
\begin{center}
\figspace
\begin{minipage}{0.36\textwidth}
\includegraphics[width=.99\textwidth]{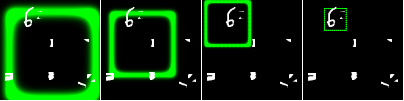}
\includegraphics[width=.99\textwidth]{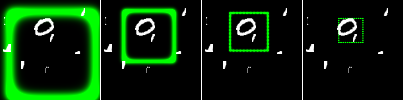}
\includegraphics[width=.99\textwidth]{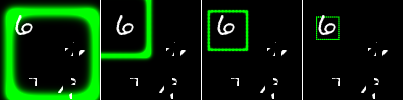}
\includegraphics[width=.99\textwidth]{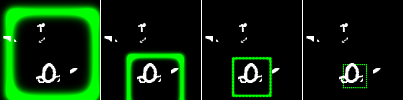}
\includegraphics[width=.99\textwidth]{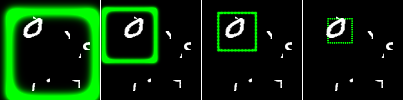}
\includegraphics[width=.99\textwidth]{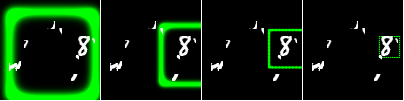}
\includegraphics[width=0.18\linewidth]{fig/time_arrow.pdf}
\end{minipage}
\end{center}
\caption{\textbf{Cluttered MNIST classification with attention.} Each sequence shows a succession of four glimpses taken by the network while classifying cluttered translated MNIST. The green rectangle indicates the size and location of the attention patch, while the line width represents the variance of the filters.
}
\label{fig:filterSum}
\end{figure}

To test the classification efficacy of the DRAW attention mechanism (as opposed to its ability to aid in image generation), we evaluate its performance on the $100 \times 100$ cluttered translated MNIST task~\cite{mnih2014recurrent}.
Each image in cluttered MNIST contains many digit-like fragments of visual clutter that the network must distinguish from the true digit to be classified.
As illustrated in Fig.~\ref{fig:filterSum}, having an iterative attention model allows the network to progressively zoom in on the relevant region of the image, and ignore the clutter outside it. 

Our model consists of an LSTM recurrent network that receives a $12 \times 12$ `glimpse' from the input image at each time-step, using the selective \emph{read} operation defined in Section~\ref{sec:attention}.
After a fixed number of glimpses the network uses a softmax layer to classify the MNIST digit.
The network is similar to the recently introduced Recurrent Attention Model (RAM) \citep{mnih2014recurrent}, except that our attention method is differentiable; we therefore refer to it as ``Differentiable RAM''.

The results in \tbl{tab:mnist100c} demonstrate a significant improvement in test error over the original RAM network.
Moreover our model had only a single attention patch at each time-step, whereas RAM used four, at different zooms.

\begin{table}[t]
\caption{Classification test error on $100 \times 100$ Cluttered Translated MNIST.}
\label{tab:mnist100c}
\begin{center}
\begin{tabular}{l|cccccccc}
\hline
Model & Error \\
\hline
Convolutional, 2 layers & 14.35\%\\
RAM, 4 glimpses, $12\times 12$, 4 scales & 9.41\%  \\
RAM, 8 glimpses, $12\times 12$, 4 scales & 8.11\%  \\
\hline
Differentiable RAM, 4 glimpses, $12\times 12$ & 4.18\% \\
Differentiable RAM, 8 glimpses, $12\times 12$ & \textbf{3.36}\% \\
\end{tabular}
\end{center}
\vskip -0.1in
\end{table}

\subsection{MNIST Generation}
We trained the full DRAW network as a generative model on the binarized MNIST dataset~\cite{salakhutdinov2008}.
This dataset has been widely studied in the literature, allowing us to compare the numerical performance (measured in average nats per image on the test set) of DRAW with existing methods.
\tbl{tab:binmnist} shows that DRAW without selective attention performs comparably to other recent generative models such as DARN, NADE and DBMs, and that DRAW with attention considerably improves on the state of the art.

\begin{table}[t]
\caption{Negative log-likelihood (in nats) per test-set example on the binarised MNIST
data set.
The right hand column, where present, gives an upper bound (Eq.~\ref{eq:loss}) on the negative log-likelihood.
The previous results are from
[1] \citep{salakhutdinov2009deep},
[2] \citep{murray2009evaluating},
[3] \citep{uria2014deep},
[4] \citep{raiko2014nadek},
[5] \citep{rezende2014stochastic},
[6] \citep{salimans2014markov},
[7] \citep{gregor2013deep}.
}
\label{tab:binmnist}
\begin{center}
\begin{tabular}{l|cccccccc}
\hline
Model & $-\log p$ & $\le$ \\
\hline
DBM 2hl [1] & $\approx 84.62$ &\\
DBN 2hl [2] & $\approx 84.55$ &\\
NADE [3] & $88.33$ & \\
EoNADE 2hl (128 orderings) [3] & $85.10$ & \\
EoNADE-5 2hl (128 orderings) [4] & $84.68$ & \\
DLGM [5] & $\approx 86.60$ &\\
DLGM 8 leapfrog steps [6] & $\approx 85.51$ & $88.30$ \\
DARN 1hl [7] & $\approx 84.13$ & $88.30$ \\
DARN 12hl [7] & - & 87.72 \\
\hline
DRAW without attention & - & 87.40 \\
DRAW & - & \textbf{80.97} \\
\end{tabular}
\end{center}
\vskip -0.1in
\end{table}

\begin{figure}[ht!]
\begin{center}
\figspace
\includegraphics[width=.45\textwidth]{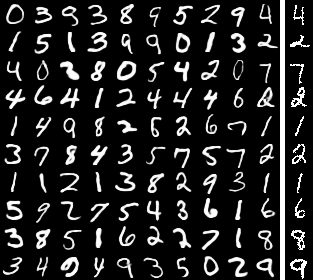}
\end{center}
\caption{\textbf{Generated MNIST images.} All digits were generated by DRAW except those in the rightmost column, which shows the training set images closest to those in the column second to the right (pixelwise $L^2$ is the distance measure). Note that the network was trained on binary samples, while the generated images are mean probabilities.
}
\label{fig:mnist_generated_nearest}
\end{figure}
\begin{figure}[ht!]
\begin{center}
\figspace
\begin{minipage}{\figwidth}
\includegraphics[width=.99\textwidth]{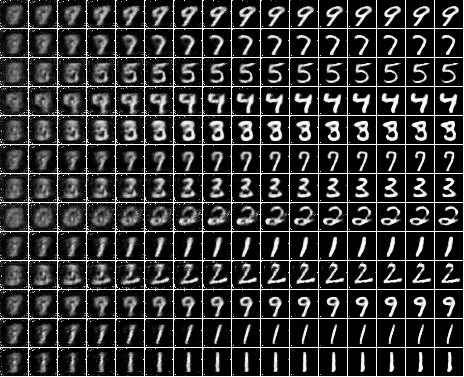}
\includegraphics[width=0.18\linewidth]{fig/time_arrow.pdf}
\end{minipage}
\end{center}
\caption{\textbf{MNIST generation sequences for DRAW without attention.} Notice how the network first generates a very blurry image that is subsequently refined.
}
\label{fig:genSeqNoAtt}
\end{figure}
Once the DRAW network was trained, we generated MNIST digits following the method in Section~\ref{sec:generation}, examples of which are presented in
Fig.~\ref{fig:mnist_generated_nearest}.
Fig.~\ref{fig:genSeqNoAtt} illustrates the image generation sequence for a DRAW network without selective attention (see Section~\ref{sec:noattention}).
It is interesting to compare this with the generation sequence for DRAW with attention, as depicted in Fig.~\ref{fig:mnist_generative_process}.
Whereas without attention it progressively sharpens a blurred image in a global way, with attention it constructs the digit by tracing the lines---much like a person with a pen.

\subsection{MNIST Generation with Two Digits}
\begin{figure}[t]
\begin{center}

\figspace

\includegraphics[width=.475\textwidth]{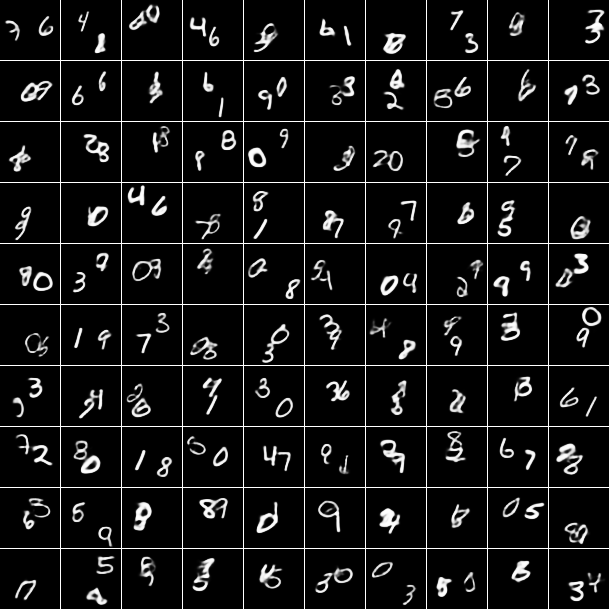}

\end{center}
\caption{\textbf{Generated MNIST images with two digits.}
}
\label{fig:TwoDigits}
\end{figure}

The main motivation for using an attention-based generative model is that large images can be built up iteratively, by adding to a small part of the image at a time.
To test this capability in a controlled fashion, we trained DRAW to generate images with two $28 \times 28$ MNIST images chosen at random and placed at random locations in a $60 \times 60$ black background. 
In cases where the two digits overlap, the pixel intensities were added together at each point and clipped to be no greater than one.
Examples of generated data are shown in Fig.~\ref{fig:TwoDigits}.
The network typically generates one digit and then the other, suggesting an ability to recreate composite scenes from simple pieces.

\subsection{Street View House Number Generation}
MNIST digits are very simplistic in terms of visual structure, and we were keen to see how well DRAW performed on natural images.
Our first natural image generation experiment used the multi-digit Street View House Numbers dataset~\citep{netzer2011reading}.
We used the same preprocessing as \citep{goodfellow2013multi}, yielding a $64 \times 64$ house number image for each training example.
The network was then trained using $54 \times 54$ patches extracted at random locations from the preprocessed images.
The SVHN training set contains 231,053 images, and the validation set contains 4,701 images.

\begin{figure}[t]
\begin{center}
\figspace
\includegraphics[width=\figwidth]{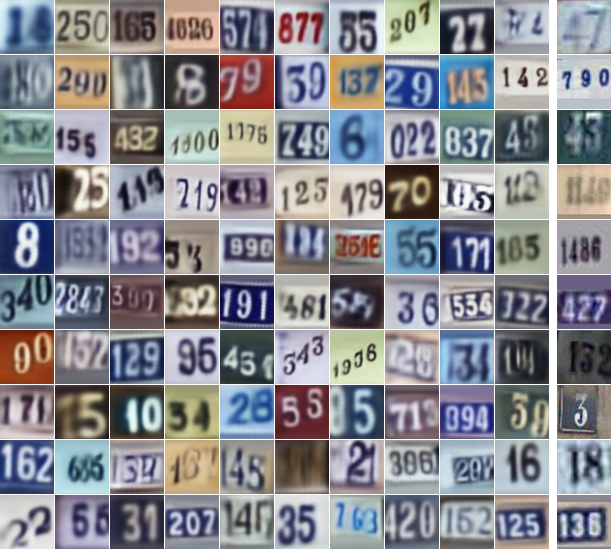}
\end{center}
\caption{\textbf{Generated SVHN images.} The rightmost column shows the training images closest (in $L^2$ distance) to the generated images beside them. Note that the two columns are visually similar, but the numbers are generally different.}
\label{fig:svhn_generated_nearest}
\end{figure}

\begin{figure}[ht!]
\figspace
\begin{center}
\begin{minipage}{\figwidth}
\includegraphics[width=0.99\textwidth]{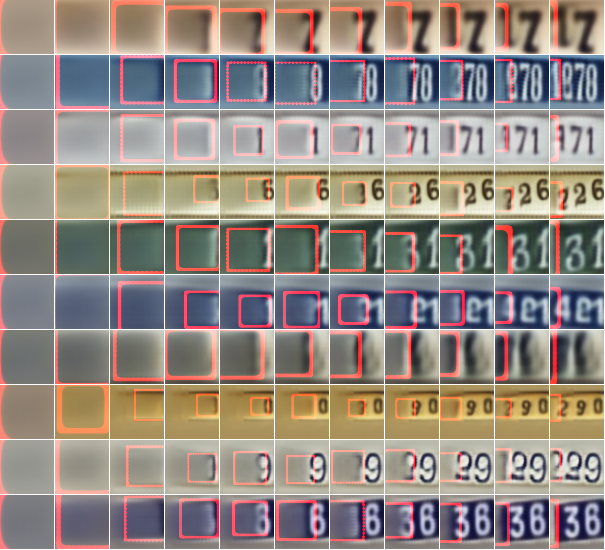}
\includegraphics[width=0.18\linewidth]{fig/time_arrow.pdf}
\end{minipage}
\end{center}
\caption{\textbf{SVHN Generation Sequences.}
The red rectangle indicates the attention patch.
Notice how the network draws the digits one at a time, and how it moves and scales the writing patch to produce numbers with different slopes and sizes.
}
\label{fig:svhn_generative_process}
\end{figure}

\begin{figure}[h!]
\begin{center}
\figspace
\includegraphics[width=\figwidth]{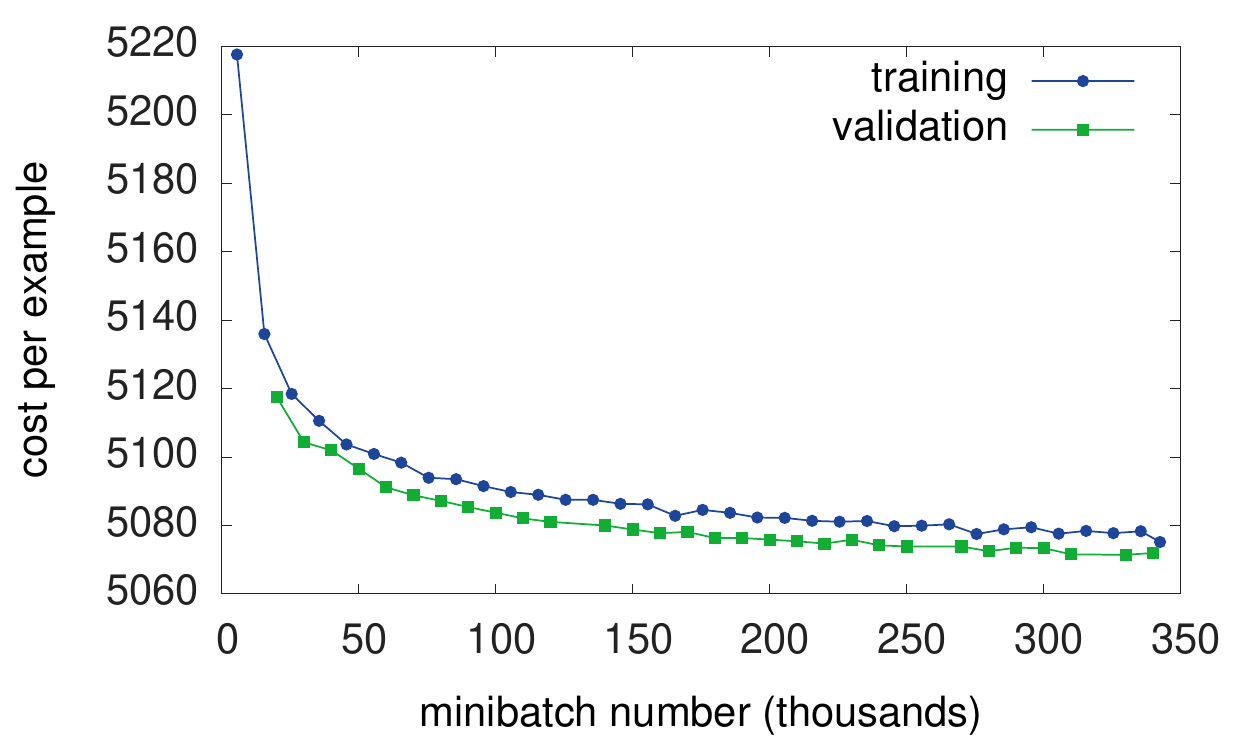}
\end{center}
\caption{\textbf{Training and validation cost on SVHN.} The validation cost is consistently lower because the validation set patches were extracted from the image centre (rather than from random locations, as in the training set). The network was never able to overfit on the training data.
}
\label{fig:svhn_cost}
\end{figure}

The house number images generated by the network are highly realistic, as shown in Figs.~\ref{fig:svhn_generated_nearest} and ~\ref{fig:svhn_generative_process}.
Fig.~\ref{fig:svhn_cost} reveals that, despite the long training time, the DRAW network underfit the SVHN training data.

\begin{table*}[t]
\caption{Experimental Hyper-Parameters.}
\label{tab:netdetails}
\begin{center}
\begin{tabular}{l r r r c c r r}
\hline
Task & \#glimpses & LSTM \#$h$ & \#$z$ & Read Size & Write Size \\
\hline
$100 \times 100$ MNIST Classification & $8$ & $256$ & - & $12 \times 12$ & - \\
MNIST Model & $64$ & $256$ & $100$ & $2 \times 2$ & $5 \times 5$ \\
SVHN Model & $32$ & $800$ & $100$ & $12 \times 12$ & $12 \times 12$ \\
CIFAR Model & $64$ & $400$ & $200$ & $5 \times 5$ & $5 \times 5$ \\
\hline
\end{tabular}
\end{center}
\vskip -0.1ins
\end{table*}

\subsection{Generating CIFAR Images}
\begin{figure}[t]
\begin{center}
\figspace
\includegraphics[width=\figwidth]{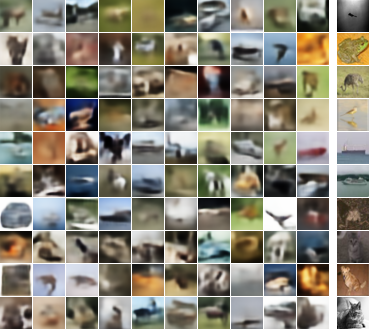}
\end{center}
\caption{\textbf{Generated CIFAR images.}
The rightmost column shows the nearest training examples to the column beside it.}
\label{fig:cifarGenerations}
\end{figure}

The most challenging dataset we applied DRAW to was the CIFAR-10 collection of natural images~\citep{krizhevsky2009learning}.
CIFAR-10 is very diverse, and with only 50,000 training examples it is very difficult to generate realistic-looking objects without overfitting (in other words, without copying from the training set).
Nonetheless the images in Fig.~\ref{fig:cifarGenerations} demonstrate that DRAW is able to capture much of the shape, colour and composition of real photographs.

\section{Conclusion}\label{sec:conclusion}
This paper introduced the Deep Recurrent Attentive Writer (DRAW) neural network architecture, and demonstrated its ability to generate highly realistic natural images such as photographs of house numbers, as well as improving on the best known results for binarized MNIST generation.
We also established that the two-dimensional differentiable attention mechanism embedded in DRAW is beneficial not only to image generation, but also to image classification.

\section*{Acknowledgments}
Of the many who assisted in creating this paper, we are especially thankful to
Koray Kavukcuoglu, Volodymyr Mnih, Jimmy Ba, Yaroslav Bulatov,
Greg Wayne, Andrei Rusu and Shakir Mohamed.

\clearpage

\normalsize{
\bibliography{draw}
}
\bibliographystyle{icml2015}

\end{document}